\documentclass[conference]{IEEEtran}

\usepackage{cite}
\usepackage{amsmath,amssymb,amsfonts}
\usepackage{algorithmic}
\usepackage{graphicx}
\usepackage{textcomp}
\usepackage{xcolor}
\usepackage{comment}
\usepackage{cleveref}
\usepackage{caption}
\usepackage{subcaption}
\usepackage{multirow} 
\usepackage{booktabs} 
\usepackage{breqn}
\usepackage{csquotes}

\usepackage{hyperref}

\begin{document}


\title{Demographic User Modeling for Social Robotics with Multimodal Pre-trained Models

}

\author{\IEEEauthorblockN{Hamed Rahimi, Mouad Abrini, Mahdi Khoramshahi, and Mohamed Chetouani}
\IEEEauthorblockA{\textit{Institut des Systèmes Intelligents et de Robotique} \\
\textit{Sorbonne University}\\
Paris, France \\
\{firstname.lastname\}@sorbonne-universite.fr}
}

\maketitle

\begin{abstract}


This paper investigates the performance of multimodal pre-trained models in user profiling tasks based on visual-linguistic demographic data. These models are critical for adapting to the needs and preferences of human users in social robotics, thereby providing personalized responses and enhancing interaction quality. First, we introduce two datasets specifically curated to represent demographic characteristics derived from user facial images. Next, we evaluate the performance of a prominent contrastive multimodal pre-trained model, CLIP, on these datasets, both in its out-of-the-box state and after fine-tuning. Initial results indicate that CLIP performs suboptimal in matching images to demographic descriptions without fine-tuning. Although fine-tuning significantly enhances its predictive capacity, the model continues to exhibit limitations in effectively generalizing subtle demographic nuances. To address this, we propose adopting a masked image modeling strategy to improve generalization and better capture subtle demographic attributes. This approach offers a pathway for enhancing demographic sensitivity in multimodal user modeling tasks.

\end{abstract}


\section{Introduction}

Developing intuitive and adaptive social robotics requires AI systems that effectively perceive individual needs, behaviors, and preferences~\cite{romeo2022exploring}. These systems can dynamically tailor their responses to enhance user engagement and satisfaction by leveraging visual-linguistic features such as demographic information~\cite{frith2005theory}. This perception-driven approach not only improves interaction quality and personalization but also addresses privacy concerns by focusing on context-sensitive observations. This is particularly important in delicate applications such as healthcare~\cite{chisty2024smart, cavallini2021can} and education~\cite{kristen2014theory}, where comprehending user behavior is key to ensuring safety and enhancing user experience. Analyzing user profiles, preferences, and behaviors is an integral component of the broader domain of user modeling tasks~\cite{purificato2024user}. This task includes methodologies that capture and represent user features and personal characteristics to create accurate user profiles. The ultimate goal of user modeling is to adapt interactions to meet users' cognitive and physical needs. Traditionally, user modeling has relied on collecting and analyzing sensitive user information, raising significant privacy concerns~\cite{zhou2020effect}.
\begin{figure}[t]
    \centering
    \includegraphics[width=\linewidth]{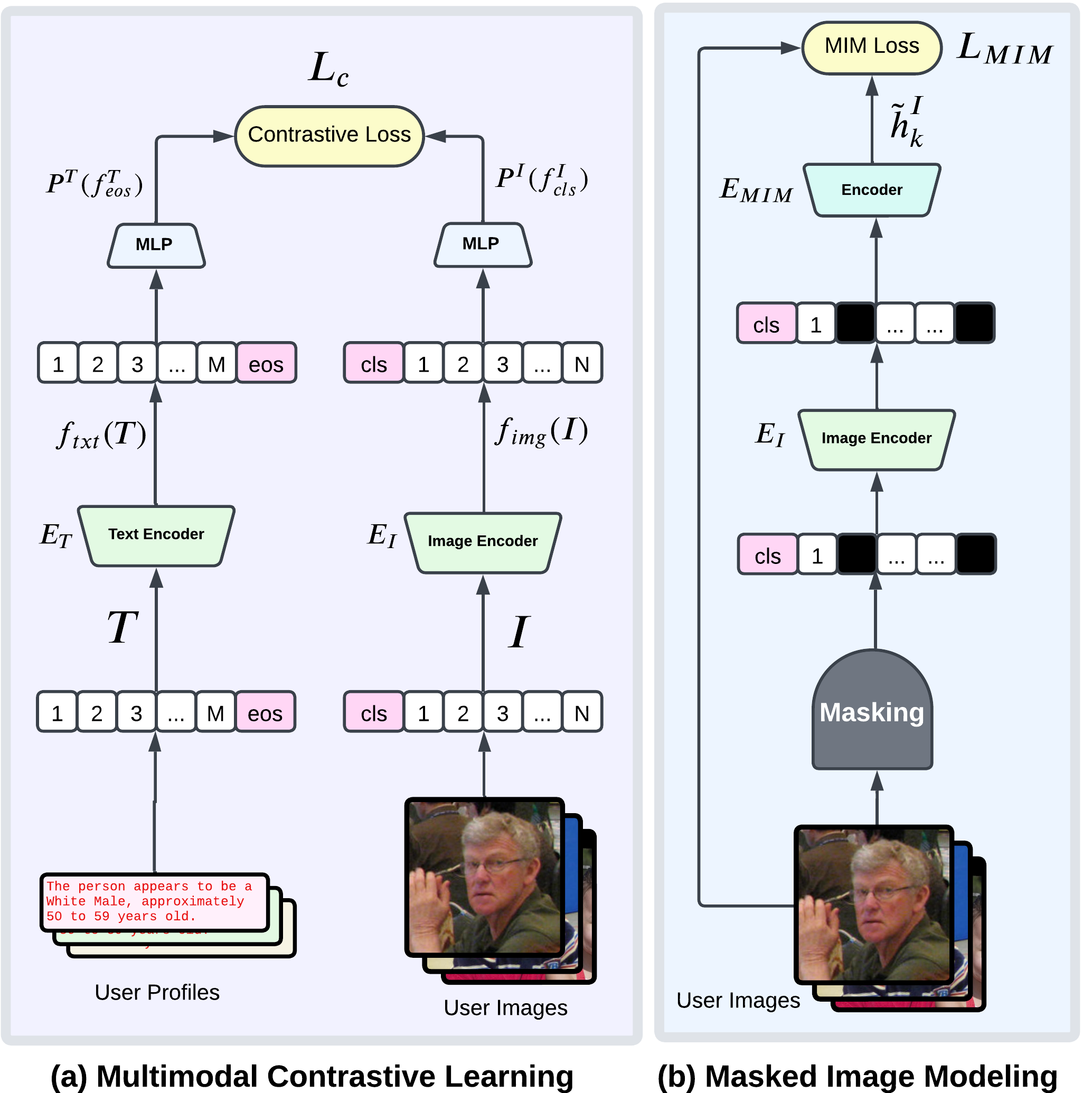}
    \caption{Proposed Demographic User Model}
    \label{fig:arch}
\end{figure}
Early approaches primarily focused on constructing personalized user profiles through historical data and expert-validated behavioral rules~\cite{adomavicius2001expert}. These methods often required extensive manual input, which posed challenges in terms of scalability and adaptability to dynamic user behaviors. 
Recent advances have shifted towards statistical and machine learning-based approaches to overcome these limitations~\cite{bao2012unsupervised}. 
While these approaches improved the handling of diverse real-world data, they struggled with adapting to new user types and unrepresented behaviors. Applications of these methods in social robotics further demonstrated the potential of neural architectures to enable memory-dependent tasks~\cite{doering2019neural}, although issues of scalability, flexibility, and parallelization persisted.

 Multimodal pre-training research~\cite{zhang2024mm} has demonstrated significant success across a spectrum of downstream tasks in recent years. This success has sparked remarkable interest in their potential to empower the next generation of personalized AI agents~\cite{xie2024large} that can learn individual needs, behaviors, and preferences. More recently, pre-trained models such as UserBERT~\cite{wu2022userbert} have gained prominence for their ability to capture interaction patterns without extensive private data storage. Building on this, innovations like User-LLM~\cite{ning2024user} have integrated user modeling with generative language models, offering a promising direction for scalable, flexible, and accurate personalized systems that also prioritize user privacy. However, these advancements are predominantly monomodal and often language-oriented, leaving the integration of multimodal data—critical for holistic user modeling—an open research challenge. Multimodal pre-trained models such as CLIP\cite{yuan2021multimodal} offer a promising foundation for extending this research to visual-linguistic user modeling, enabling richer and more comprehensive interaction paradigms.

 However, the sparse and sensitive nature of visual-linguistic demographic data, including attributes such as gender, age, and ethnicity, poses significant challenges for multimodal pre-trained models like CLIP, which rely on contrastive training, to perform user modeling and profile representation tasks. To address these limitations, we propose integrating contrastive learning with masked image modeling (MIM), as shown in \Cref{fig:arch}, where multimodal pre-trained models are encouraged to predict occluded regions in images during training. This approach could enhance the model's ability to capture nuanced demographic patterns while mitigating biases.

 This paper contributes to the literature on multimodal user modeling by (1) introducing two novel datasets designed to capture diverse visual-linguistic demographic traits for user profiling; (2) evaluating the performance of state-of-the-art multimodal models both in their out-of-the-box configuration and after fine-tuning, on these datasets; and (3) present an approach to improve generalization by leveraging masked image modeling techniques, which enhances the model's ability to capture subtle demographic attributes effectively.



\begin{figure*}[htbp]
  \centering
  \begin{subfigure}[b]{0.32\textwidth}
    \includegraphics[width=\textwidth]{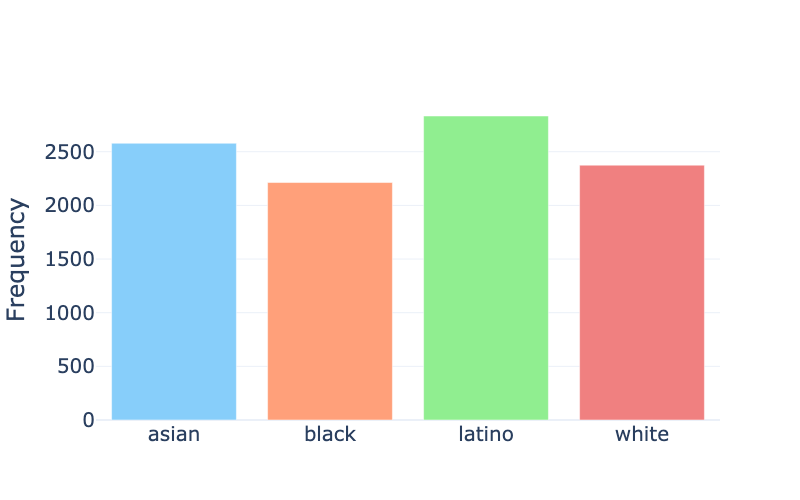}
    \caption{Ethnicity}
  
  \end{subfigure}
  \hfill
    \begin{subfigure}[b]{0.32\textwidth}
    \includegraphics[width=\textwidth]{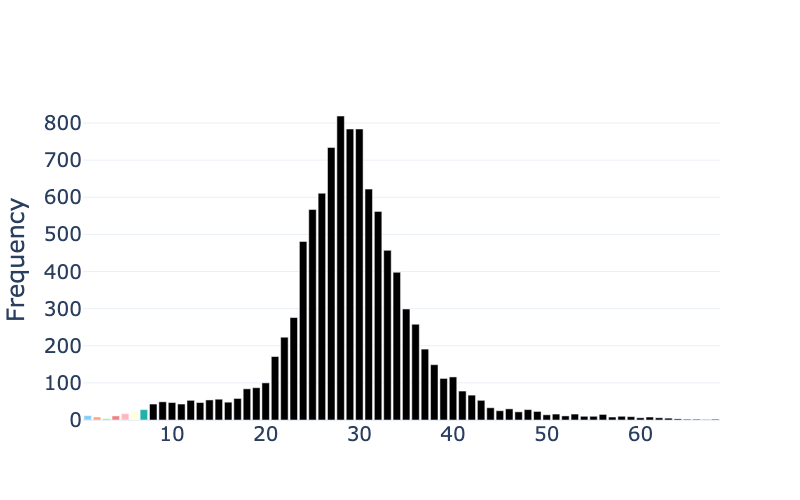}
    \caption{Age}

  \end{subfigure}
  \hfill
  \begin{subfigure}[b]{0.32\textwidth}
    \includegraphics[width=\textwidth]{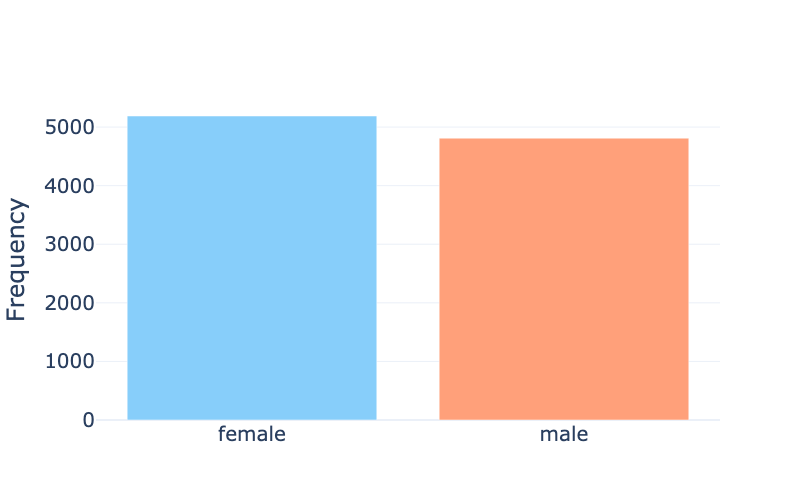}
    \caption{Gender}

  \end{subfigure}
\caption{Distribution of Age, Gender, and Ethnicity of entries in GenUser}
\label{fig:dis}
\end{figure*}

\section{Background}
\label{sec:lit}

Pre-trained models have been widely explored for user modeling, with notable advancements such as Pre-training User Model (PTUM)~\cite{wu2020ptum} and UserBERT~\cite{wu2022userbert}. Both models leverage large-scale sequential user behavior to represent user models in a high-dimensional space. Despite the significant progress in pre-trained behavior-based user models, there has been limited exploration of multimodal pre-trained models specifically designed for demographic user modeling based on extracting facial features. However, recent advancements in multimodal contrastive pre-trained models offer new potential for user modeling by integrating diverse data types, such as visual and textual cues, enabling more nuanced analysis of facial features alongside behavioral data~\cite{liu2022multi}. 

Multimodal contrastive pre-trained models such as CLIP~\cite{yuan2021multimodal} are designed to learn joint representations from various data modalities, primarily visual and textual information. They achieve this by ensuring that similar instances are represented closely in the space while dissimilar instances are kept far apart~\cite{chopra2005learning}. As shown in \Cref{fig:arch}(a), this process operates on a dataset $\mathcal{D} = \{(I_i, T_i)\}_{i=1}^N$ comprising $N$ paired instances of images and text, where each image $I_i \in \mathbb{R}^{d_I \times N}$ and text $T_i \in \mathbb{R}^{d_T \times M}$ are represented as sequences of tokens. The indices $N$ and $M$ denote the sequential lengths of image and text tokens, respectively, while $d_I$ and $d_T$ represent their corresponding feature dimensions. The framework employs two primary encoder functions: an image encoder $E_I: \mathbb{R}^{d_I \times N} \rightarrow \mathbb{R}^{d_h \times N}$ and a text encoder $E_T: \mathbb{R}^{d_T \times M} \rightarrow \mathbb{R}^{d_h \times M}$, where $d_h$ denotes the hidden dimension. These Transformer-based encoders process their respective modalities to produce sequences of feature vectors $E_I(I) = \{f^I_1, f^I_2, \ldots, f^I_N\}$ and $E_T(T) = \{f^T_1, f^T_2, \ldots, f^T_M\}$. Special tokens, namely \texttt{cls} for images and \texttt{eos} for text, are incorporated into these sequences to capture global representations. A projection head $P: \mathbb{R}^{d_h} \rightarrow \mathbb{R}^{d_e}$, implemented as a multilayer perceptron, maps the features associated with these special tokens to a shared embedding space, yielding $e_I = P(f^I_{\text{cls}})$ for images and $e_T = P(f^T_{\text{eos}})$ for text. The similarity between image and text embeddings is computed using a similarity function $s_{i,j} = \text{sim}(e_I^i, e_T^j)$, typically implemented as cosine similarity. The learning objective is formalized through a bidirectional contrastive loss function $\mathcal{L}_{\text{c}}$, defined as:

\begin{dmath}
\label{equation:contt}
  \mathcal{L}_{\text{c}} = -\frac{1}{2N} \Big( \sum_{i=1}^N \Big( \log p\Big(f_{img}(I_i)| f_{txt}(T_i) \Big )
    \nonumber \\
    + \log p\Big(f_{txt}(T_i) | f_{img}(I_i)\Big ) \Big ) \Big)
\end{dmath}
where the matching probability for any image-text pair is computed using a softmax function with temperature parameter $\tau$:
\begin{equation}
p\big(f_{\text{txt}}(T_i)| f_{\text{img}}(I_i)\big) = \frac{\exp\left(\frac{s_{i,i}}{\tau}\right)}{\sum_{k=1}^N \exp\left(\frac{s_{i,k}}{\tau}\right)}
\end{equation}
\begin{equation}
p(f_{\text{img}}(I_i) | f_{txt}(T_i)) = \frac{\exp\left(\frac{s_{i,i}}{\tau}\right)}{\sum_{k=1}^N \exp\left(\frac{s_{k,i}}{\tau}\right)}
\end{equation}
This formulation ensures bidirectional alignment between modalities by simultaneously optimizing the probability of matching images to their corresponding texts and vice versa. The temperature parameter $\tau$ modulates the distribution of similarities during training. Through minimization of this contrastive loss, the framework learns to embed both visual and textual content in a shared semantic space where matched pairs are proximal while unmatched pairs are distant.

Recent studies by AdaptSSR~\cite{yu2024adaptssr} and FairLISA~\cite{liu2024fairlisa} have highlighted inherent limitations of these models in user modeling with sensitive demographic data such as gender, which underscore gaps in fairness and adaptability within such models. AdaptSSR~\cite{yu2024adaptssr} addresses data sparsity issues by introducing Augmentation-Adaptive Self-Supervised Ranking, while FairLISA~\cite{liu2024fairlisa} tackles bias arising from sensitive attributes using an adversarial framework. In this paper, we build on a general facial Representation model (FaRL~\cite{zheng2022general}) that is trained on LAION-FACE (a subset of the LAION dataset~\cite{schuhmann2022laion}) using both visual and textual modalities to develop robust multimodal facial representations. 
We apply this technique to our problem and investigate the impact of integrating the masked image modeling (MIM) objective with a multimodal contrastive pre-trained model for demographic user modeling tasks.


\section{Methods}
\label{sec:method}
\subsection{Dataset Construction}
\label{sec:data}
We introduce two datasets designed to capture diverse visual-linguistic demographic traits for user profiling.

\subsubsection{GenUser}

\begin{figure}[h]
    \centering
    \includegraphics[width=\linewidth]{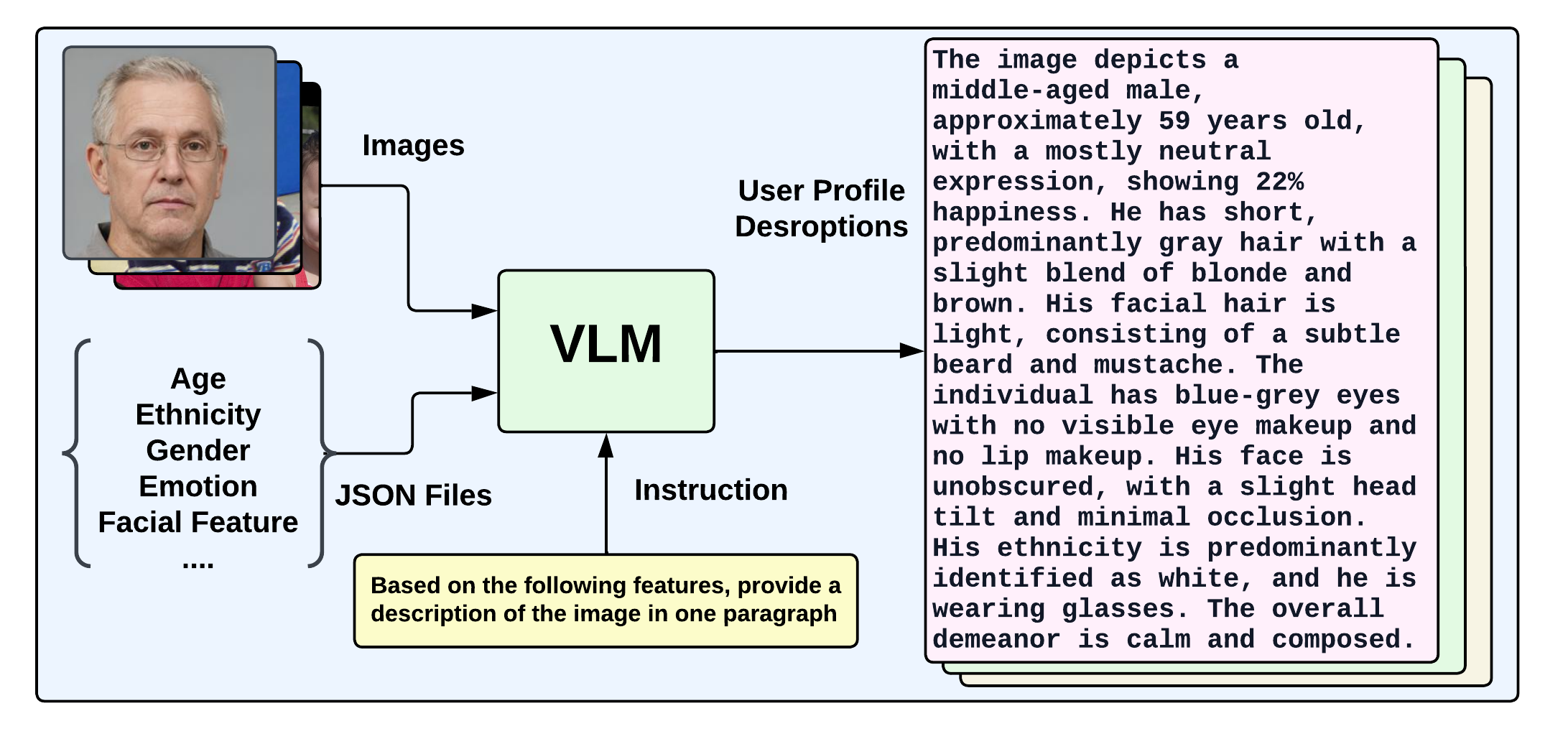}
    \caption{GenUser Data Construction Pipeline}
    \label{fig:genuser}
\end{figure}

It includes 10K synthetic image-text pairs, featuring human faces alongside user profile information from diverse demographic backgrounds. The dataset is generated by \enquote{\textit{generated.photos}} platform to ensure privacy and avoid using real personal data. To promote fairness, as shown in \Cref{fig:dis}, 
the entries are intentionally designed to represent a broad range of demographic groups, capturing diversity across key characteristics such as age, gender, and ethnicity. 
Each entry is accompanied by a JSON file integrating over 10 visual attributes that support a wide range of information about user profiles.
As shown in \Cref{fig:genuser}, these features, alongside the images, are processed using a VLM (\enquote{GPT-4o}) to generate a one-paragraph user profile, providing a concise yet detailed description based on the inferred demographic and emotional attributes. The 10K entries in the dataset are split into three parts: 1K for validation, 1K for testing, and 8K for training, ensuring a balanced distribution across the dataset for training and model evaluation.

\subsubsection{FairUser}
It consists of 100K real-world text-image pairs derived from the FairFace dataset~\cite{karkkainen2021fairface}. The dataset entries are carefully curated to ensure balance, diversity, and accurate labeling across race, gender, and age categories. Based on this dataset, we designed a user profile feature using the following template: \enquote{The person appears to be {race class} {gender class}, approximately {age class} years old}. This template facilitates a structured and interpretable representation of demographic attributes for profiling tasks. The 100K entries in the dataset are split into three parts: 10K for validation, 10K for testing, and 80K for training, ensuring a balanced distribution across the dataset for training and model evaluation.


\begin{table*}[hbtp]
\centering
\caption{Performance Comparison for User Retrieval Task}
\label{tab:performance}
\begin{tabular}{cccccccccc}
\toprule
\multirow{3}{*}{Test Set} & \multirow{3}{*}{Metric}        & \multicolumn{6}{c}{Multimodal Pre-trained Model}   \\ \cmidrule(r){3-8}
&   & \multicolumn{2}{c}{Out-of-the-box} & \multicolumn{2}{c}{Finetuned on GenUser} & \multicolumn{2}{c}{Finetuned on FairUser} \\ \cmidrule(r){3-4} \cmidrule(r){5-6} \cmidrule(r){7-8}
& & CLIP   & \multicolumn{1}{c}{FaRL}  & CLIP      & \multicolumn{1}{c}{FaRL}     & CLIP      & \multicolumn{1}{c}{FaRL}     \\ 
\midrule\multirow{4}{*}{\rotatebox{90}{GenUser}}  
                          & Accuracy  & 0.161            & 0.133           & 0.668 \textcolor{green}{$\uparrow$}             & 0.659  \textcolor{green}{$\uparrow$}           & 0.145 \textcolor{red}{$\downarrow$}               & 0.206   \textcolor{green}{$\uparrow$}            \\
                          & Recall    & 0.160            & 0.132           & 0.667 \textcolor{green}{$\uparrow$}             & 0.658   \textcolor{green}{$\uparrow$}           & 0.144   \textcolor{red}{$\downarrow$}            & 0.205   \textcolor{green}{$\uparrow$}             \\
                          & Precision & 0.179            & 0.138           & 0.680 \textcolor{green}{$\uparrow$}             & 0.662  \textcolor{green}{$\uparrow$}            & 0.151  \textcolor{red}{$\downarrow$}             & 0.218   \textcolor{green}{$\uparrow$}             \\
                          & F1        & 0.162            & 0.129           & 0.668 \textcolor{green}{$\uparrow$}             & 0.656   \textcolor{green}{$\uparrow$}            & 0.142    \textcolor{red}{$\downarrow$}           & 0.203  \textcolor{green}{$\uparrow$}             \\ \midrule
\multirow{4}{*}{\rotatebox{90}{FairUser}} 
                            & Accuracy  & 0.345            & 0.236           & 0.291   \textcolor{red}{$\downarrow$}             & 0.269   \textcolor{green}{$\uparrow$}            & 0.493   \textcolor{green}{$\uparrow$}             & 0.516   \textcolor{green}{$\uparrow$}            \\
                          & Recall    & 0.345            & 0.235           & 0.291     \textcolor{red}{$\downarrow$}           & 0.269  \textcolor{green}{$\uparrow$}            & 0.492 \textcolor{green}{$\uparrow$}               & 0.516   \textcolor{green}{$\uparrow$}             \\
                          & Precision & 0.350            & 0.239           & 0.297     \textcolor{red}{$\downarrow$}           & 0.272  \textcolor{green}{$\uparrow$}           & 0.505    \textcolor{green}{$\uparrow$}            & 0.530   \textcolor{green}{$\uparrow$}             \\
                          & F1        & 0.342            & 0.234           & 0.289   \textcolor{red}{$\downarrow$}           & 0.267  \textcolor{green}{$\uparrow$}             & 0.491     \textcolor{green}{$\uparrow$}           & 0.514   \textcolor{green}{$\uparrow$}             \\  \bottomrule

\end{tabular}
\end{table*}

\subsection{Demographic User Modeling}

As illustrated in \Cref{fig:arch}, let $\mathcal{D} = \{(I_i, T_i)\}_{i=1}^N$ be the training dataset of facial images and corresponding demographic descriptions. The proposed model employs two key components similar to FaRL~\cite{zheng2022general}: an image-text encoder with contrastive loss, and a masked image modeling task with cross-entropy loss. The contrastive loss is as explained in \Cref{sec:lit}, while for masked image modeling we mask some image patches in the input and predict the visual tokens corresponding to the masked patches using the same image encoder. 

Formally, let \(\tilde{I}\) be the masked image, where some image patches are randomly masked. We extract the features from the image encoder, denoted as \(\{\tilde{f}^I_{\text{cls}}, \tilde{f}_1^I, \ldots, \tilde{f}^I_N\} = E_I(I)\). These features are then fed into a small Transformer encoder, which outputs the final hidden vectors, represented as \(\{\tilde{h}_{\text{cls}}^I, \tilde{h}_1^I, \ldots, \tilde{h}_N^I\} = E_{\text{MIM}}(\tilde{f}_{\text{cls}}^I, \tilde{f}_1^I, \ldots, \tilde{f}_N^I)\). The objective is to predict the masked regions using the corresponding hidden vectors \(\{\tilde{h}_k^I : k \in M\}\). Instead of directly predicting pixel values, which would require substantial memory consumption, a discrete variational autoencoder is used to first encode each image patch to one of \(|V|\) possible values, where \(V\) is the vocabulary of the autoencoder. A classification layer is then attached to the hidden vector \(\tilde{h}_k^I\) to predict the index of the corresponding masked patch among \(\{1, \ldots, |V|\}\). The masked image modeling loss, denoted as \(\mathcal{L}_{\text{MIM}}\), can be expressed mathematically as:
\begin{equation}
\label{equ:mim}
\mathcal{L}_{\text{MIM}} = - \sum_{k \in M} \log p(q^k_\phi(I)) \mid \tilde{I}),
\end{equation}
where the conditional probability \( p(q^k_\phi(I) \mid \tilde{I}) \) represents the model's ability to predict the tokenized masked patch \( q_\phi(I) \) based on the corrupted input image \( \tilde{I} \). By incorporating this loss term, the model enhances its capacity to extract fine-grained facial characteristics while maintaining the broader semantic understanding achieved through contrastive learning. Therefore, the final loss function can be formulated as:
\begin{equation}
\mathcal{L}_{\text{total}} = \mathcal{L}_{\text{c}} + \mathcal{L}_{\text{MIM}},
\end{equation}
combining the contrastive loss \(\mathcal{L}_{\text{c}}\) in Equation (\ref{equation:contt}) with the masked image modeling loss \(\mathcal{L}_{\text{MIM}}\) in \Cref{equ:mim}.

\begin{figure*}[hbtp]
  \centering
  \begin{subfigure}[b]{0.24\textwidth}
    \includegraphics[width=\textwidth]{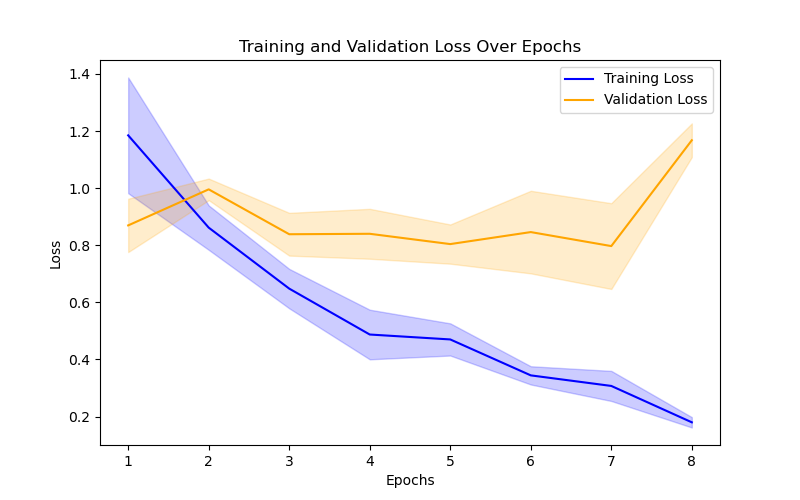}
    \caption{CLIP | GenUser}
  
  \end{subfigure}
  \hfill
    \begin{subfigure}[b]{0.24\textwidth}
    \includegraphics[width=\textwidth]{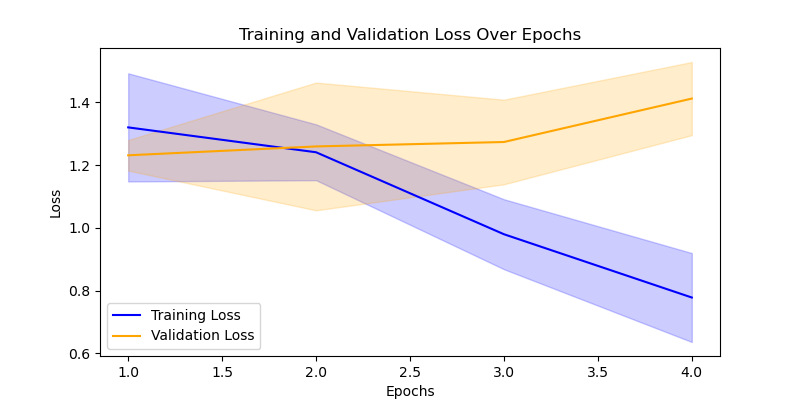}
    \caption{CLIP | FairUser}

  \end{subfigure}
  \hfill
  \begin{subfigure}[b]{0.24\textwidth}
    \includegraphics[width=\textwidth]{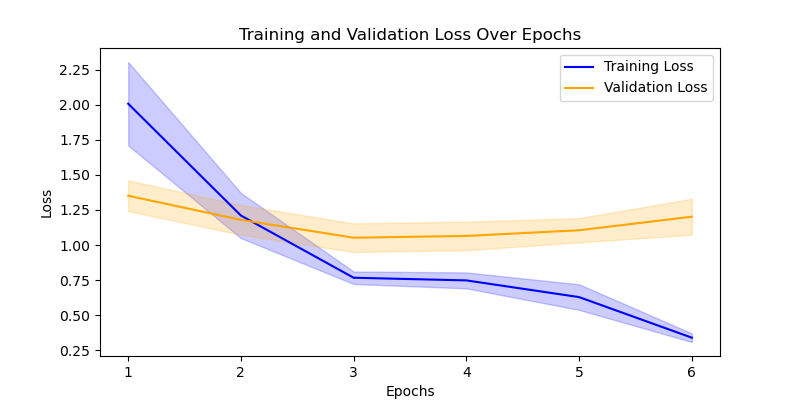}
    \caption{FaRL | GenUser}

  \end{subfigure}
  \hfill
  \begin{subfigure}[b]{0.24\textwidth}
    \includegraphics[width=\textwidth]{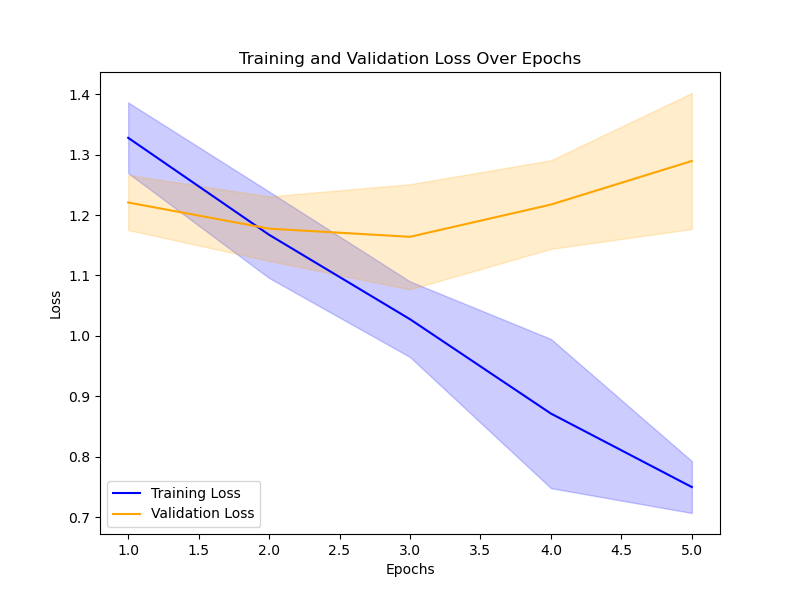}
    \caption{FaRL | FairUser}

  \end{subfigure}
\caption{Training and Validation Loss}
\label{fig:loss}
\end{figure*}

\section{Experimental Setup} 
\label{sec:exp}

\subsection{Models}

Our baseline multimodal contrastive pre-trained model is CLIP (\enquote{CLIP-ViT-Base-Patch16}), known for its robust image-text alignment capability, achieved through a contrastive loss. For our proposed model, we built on FaRL (\enquote{FaRL-Base-Patch16-ep64}) as it combines multimodal contrastive pre-training with a self-supervised learning component to enhance feature extraction from sparse or unstructured data. We compare the performance of these models both out-of-the-box and after fine-tuning to assess their inherent abilities and adaptability for user modeling tasks. 

\subsection{Finetuning Strategy}
The models are fine-tuned using a contrastive fine-tuning strategy called FLYP~\cite{goyal2023finetune}. This approach aligns the fine-tuning objective with pre-training by casting downstream class labels as text prompts and optimizing the contrastive loss between image embeddings and class-descriptive prompts. This method allows our models—CLIP and FaRL—to retain consistency with their original pre-training objectives. This strategy has shown substantial gains in few-shot learning, distribution shifts, and transfer learning benchmarks~\cite{goyal2023finetune}, making it suitable for our user profile modeling task. We set a training schedule of 10 epochs, with continuous validation to monitor performance. A stopping criterion is implemented, allowing training to halt early if validation performance does not improve for three consecutive evaluations. This early stopping threshold of three epochs helps to retain the most generalizable model parameters. The optimizer used in this study is AdamW, a variant of the Adam optimizer that improves weight decay implementation by decoupling it from the gradient update. The learning rate is set to \( 10^{-5} \). We run the models five times with different seeds on the datasets, and after fine-tuning and cross-validation, we evaluate their performance on the held-out test set to ensure exposure to previously unseen data for an unbiased assessment. 

\subsection{Metrics}
To assess the performance of these models in capturing visual attributes of user profiles, we employed four key metrics within a user retrieval task. Specifically, for each text query \(T\) in the test set, the model computes similarity scores with all images in the test set and identifies the image with the highest similarity. Accuracy is then determined by checking if this retrieved image matches the ground truth image paired with the query \(T\).

\begin{itemize} 
\item \textbf{Accuracy:} Evaluates the overall success rate of retrieving the exact correct images based on text and image embeddings. 

\item \textbf{Recall:} Evaluates the model's ability to retrieve relevant images by calculating the proportion of true positive matches out of all relevant images.

\item \textbf{Precision:} Assesses the proportion of relevant images retrieved among all retrieved images. 

\item \textbf{F1 Score:} Provides a balanced metric that combines precision and recall, particularly useful for evaluating the retrieval task when precision and recall are of equal importance. 
\end{itemize}
These metrics allow us to comprehensively evaluate each model's effectiveness in multimodal user profile retrieval, particularly in how well they capture and reflect key visual attributes in a diverse set of user images.

\section{Results}
\label{sec:res}

\subsection{Out-of-the-box Performance}
As shown in \Cref{tab:performance}, the performance of CLIP and FaRL models is relatively low in the user retrieval task across both datasets. Specifically, CLIP demonstrates a slight edge over FaRL, achieving 16\% accuracy on the GenUser dataset, which is only 3\% higher than FaRL. On the FairUser dataset, CLIP attains 34\% accuracy, outperforming FaRL by 11\%.  These results do not suggest that CLIP and FaRL models are inherently incapable of distinguishing between demographic user data. Instead, they highlight the models' challenges in accurately associating images with user profiles when processing complex queries involving sensitive and sparse features such as age, gender, ethnicity, facial attributes, and expressions.

\subsection{Fine-tuned Performance}

As shown in \Cref{fig:loss}, during fine-tuning on the GenUser dataset, the CLIP model begins to overfit after 5 epochs, while FaRL shows signs of overfitting after 4 epochs. In response, the cross-validation and thresholding mechanism halts the fine-tuning process until three additional epochs have been completed, preventing further overfitting. Despite this limitation, the average performance of both models over 5 runs improves, with CLIP achieving an F1 accuracy of 66\% and FaRL reaching 65\%. Similarly, on the FairUser dataset, CLIP begins overfitting after just 1 epoch, and FaRL after 2 epochs; here again, the cross-validation and threshold mechanism halts the training process for an additional three epochs to control overfitting. Under these conditions, the models attain F1 scores of 49\% and 51\% for CLIP and FaRL, respectively. These results demonstrate incremental performance improvements, although both models exhibit susceptibility to overfitting early in the fine-tuning process. 

It is significant to emphasize that fine-tuning FaRL on one dataset results in a performance improvement on the other dataset. This cross-dataset enhancement indicates that our assumption on the integration of masked image modeling with contrastive learning leads to a degree of generalizability, potentially benefiting from the knowledge transfer acquired during fine-tuning. In contrast, CLIP exhibits the opposite behavior: its performance on the alternative dataset declines when fine-tuned on a specific dataset. 
This behavior might stem from the significant distribution shift between the datasets used in this evaluation and those CLIP was originally trained on. As a result, CLIP essentially behaves as if it is being trained from scratch, suffering from catastrophic forgetting. The model is prone to overfitting due to the relatively small size of the datasets compared to its massive parameter count (200M), making it less adaptable to diverse or unseen data distributions.

\begin{table}[hbtp]
\centering
\caption{Example for User Retrieval Task. \textit{The query is \enquote{Image of a 17-year-old black female with freckles, wavy auburn hair in a ponytail, and green eyes, looking thoughtful with a slight smile.}}
}
\begin{tabular}{cc|c|c|c}
\toprule
 & \multicolumn{4}{c}{Multimodal Pre-trained Models}                             \\ \cmidrule(r){2-5}
 & \multicolumn{2}{c}{Out-of-the-Box} & \multicolumn{2}{c}{Finetuned on GenUser} \\ \cmidrule(r){2-3} \cmidrule(r){4-5}
 & CLIP             & FaRL            & CLIP                & FaRL               \\ \midrule
 &  \rotatebox{270}{\includegraphics[width=0.4\textwidth]{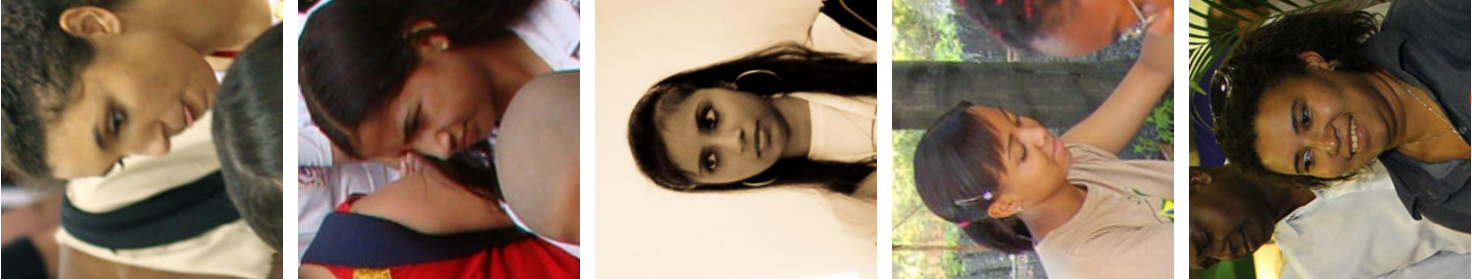}}         
 &  \rotatebox{270}{\includegraphics[width=0.4\textwidth]{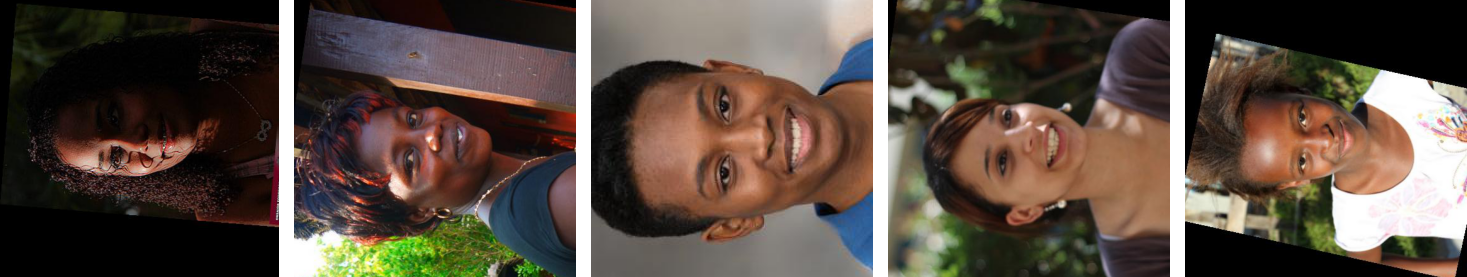}}         
 &  \rotatebox{270}{\includegraphics[width=0.4\textwidth]{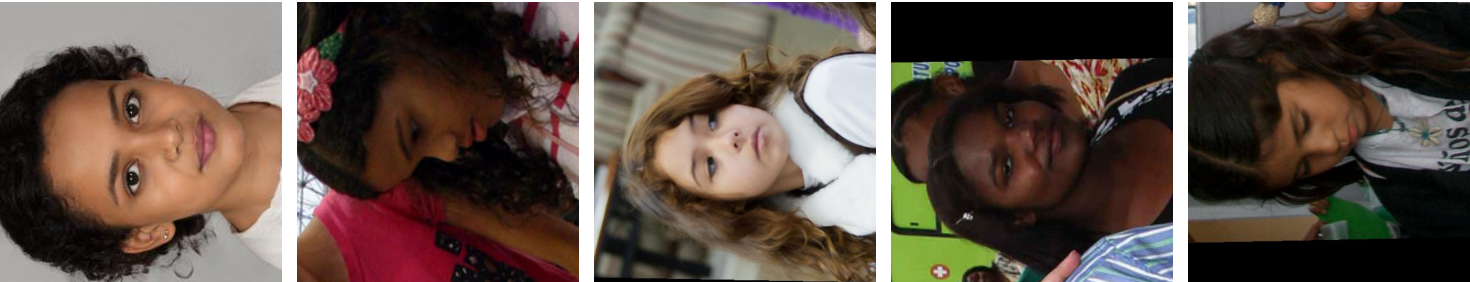}}            
 &  \rotatebox{270}{\includegraphics[width=0.4\textwidth]{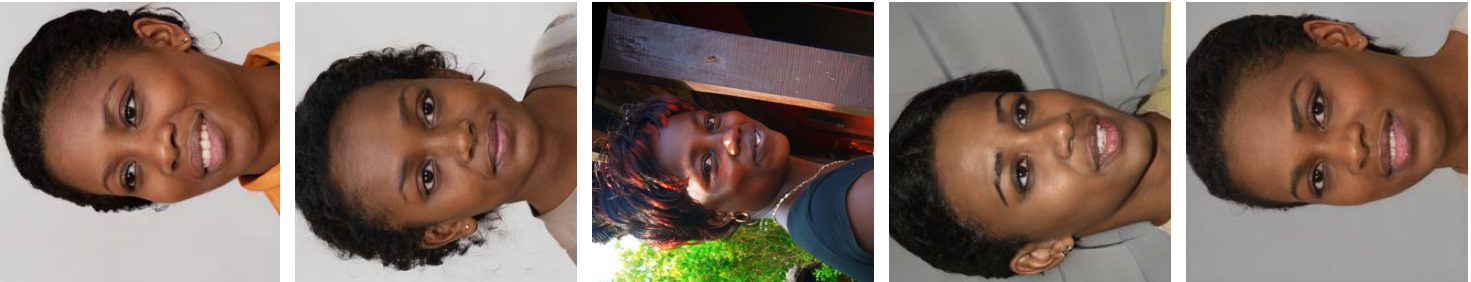}}  \\  \bottomrule      
\end{tabular}
\label{tab:exm}
\end{table}

\Cref{tab:exm} demonstrates a comparative example for the user retrieval task between CLIP and FaRL when processing a complex demographic query prompt. While CLIP successfully captures broad demographic attributes, it struggles to combine multiple specific demographic descriptors effectively. In contrast, the FaRL performs better in retrieving images that match the detailed demographic query, despite these exact attribute combinations not being present in its training data. This example highlights how general-purpose vision-language models like CLIP, though competent at basic demographic information retrieval, may have limitations when handling highly specific, multi-attribute demographic queries that require a more nuanced understanding of facial features.

\section{Conclusion}
\label{sec:conclusion}
This paper explored the effectiveness of multimodal pre-trained models in capturing demographic representations from facial features for user profiling applications. Our empirical results demonstrate that while these models perform suboptimally out-of-the-box, fine-tuning significantly enhances their predictive capacity. However, they still exhibit limitations in effectively generalizing demographic nuances. We suggest integrating a masked image modeling strategy to improve the model's generalizability and capture subtle demographic attributes. Future work will focus on refining these models and exploring additional strategies to enhance their performance in diverse and sensitive applications, such as healthcare, where understanding user behavior and preferences is critical.

\section*{Acknowledgments}

The authors sincerely acknowledge the financial support of the French National Research Agency (ANR) for the ANITA project (Grant No. ANR-22-CE38-0012-01). We also extend our gratitude to \href{https://generated.photos}{Generated Photos} for generously providing 10,000 generated face entries.

\bibliographystyle{ieeetr}

\bibliography{ref}

\end{document}